# Capsule Deep Neural Network for Recognition of Historical Graffiti Handwriting

Nikita Gordienko
National Technical University of Ukraine "Igor Sikorsky Kyiv Polytechnic Institute"
Kyiv, Ukraine
nik.gordiienko@gmail.com

Yuriy Kochura
National Technical University of Ukraine "Igor Sikorsky Kyiv Polytechnic Institute"
Kyiv, Ukraine
iuriy.kochura@gmail.com

Vlad Taran
National Technical University of Ukraine "Igor Sikorsky Kyiv Polytechnic Institute"
Kyiv, Ukraine
vladtkv@gmail.com

Peng Gang
School of Information Science and Technology, Huizhou University
Huizhou, China
peng@hzu.edu.cn

Yuri Gordienko
National Technical University of Ukraine "Igor Sikorsky Kyiv Polytechnic Institute"
Kyiv, Ukraine
yuri.gordienko@gmail.com

Sergii Stirenko
National Technical University of Ukraine "Igor Sikorsky Kyiv Polytechnic Institute"
Kyiv, Ukraine
sergii.stirenko@gmail.com

*Abstract*— Automatic recognition of the historical letters (XI-XVIII centuries) carved on the stoned walls of St.Sophia cathedral in Kyiv (Ukraine) was demonstrated by means of capsule deep learning neural network. It was applied to the image dataset of the carved Glagolitic and Cyrillic letters (CGCL), which was assembled and pre-processed recently for recognition and prediction by machine learning methods. CGCL dataset contains >4000 images for glyphs of 34 letters which are hardly recognized by experts even in contrast to notMNIST dataset with the better images of 10 letters taken from different fonts. The capsule network was applied for both datasets in three regimes: without data augmentation, with lossless data augmentation, and lossy data augmentation. Despite the much worse quality of CGCL dataset and extremely low number of samples (in comparison to notMNIST dataset) the capsule network model demonstrated much better results than the previously used convolutional neural network (CNN). The training rate for capsule network model was 5-6 times higher than for CNN. The validation accuracy (and validation loss) was higher (lower) for capsule network model than for CNN without data augmentation even. The area under curve (AUC) values for receiver operating characteristic (ROC) were also higher for the capsule network model than for CNN model: 0.88-0.93 (capsule network) and 0.50 (CNN) without data augmentation, 0.91-0.95 (capsule network) and 0.51 (CNN) with lossless data augmentation, and similar results of 0.91-0.93 (capsule network) and 0.9 (CNN) in the regime of lossless data augmentation only. The confusion matrixes were much better for capsule network than for CNN model and gave the much lower type I (false positive) and type II (false negative) values in all three regimes of data augmentation. These results supports the previous claims that capsule-like networks allow to reduce error rates not only on MNIST digit dataset, but on the other notMNIST letter dataset and the more complex CGCL handwriting graffiti letter dataset also. Moreover, capsule-like networks allow to reduce training set sizes to 180 images even like in this work, and they are considerably better than CNNs on the highly distorted and incomplete letters even like CGCL handwriting graffiti.

## I. Introduction (*Heading 1*)

The different writing traditions and techniques were invented and used by people. They were based on the available tools, carriers, and aims. The notion of graffiti applied to various writings found on the walls of modern constructions or ancient artifacts. Nowadays 'graffiti' means often any graphics applied to any surfaces and it is overused in the context of vandalism especially [1]. From the historical point of view graffiti are very powerful source of historical information. One of the most impressive examples is related to the Safaitic language which is known because the graffiti inscriptions on the surface of rocks in southern Syria, eastern Jordan and northern Saudi Arabia, are the only known sources of it [2]. But the tradition of writing graffiti is widely known on other historical places. For example, the most interesting and original examples of the Eastern Slavic writings which are preserved in their original state of medieval graffiti as visual texts are located in St. Sophia Cathedral of Kyiv (Ukraine) (Fig.1) [3].

Actually they are written in two alphabets, Glagolitic and

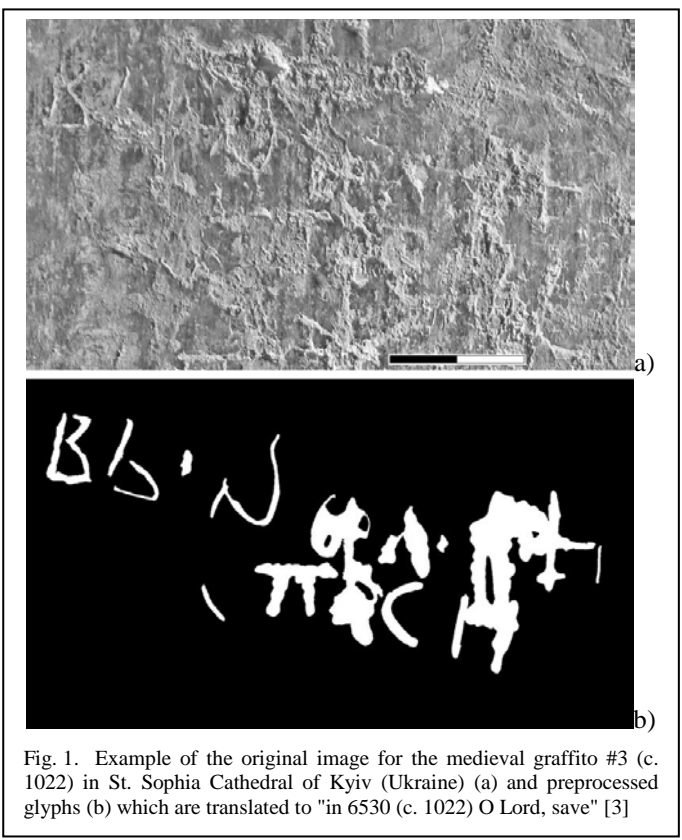

Fig. 1. Example of the original image for the medieval graffito #3 (c. 1022) in St. Sophia Cathedral of Kyiv (Ukraine) (a) and preprocessed glyphs (b) which are translated to "in 6530 (c. 1022) O Lord, save" [3]

The work was partially supported by Huizhou Science and Technology Bureau and Huizhou University (Huizhou, P.R.China) in the framework of Platform Construction for China-Ukraine Hi-Tech Park Project.

Cyrillic, and differ by the letter style, arrangement and layout (including numerous distortions like rotations of letters and incomplete form of their writing) [4,5]. The different interpretations of these graffiti were proposed by scientists as to their date, language, authorship, genuineness, and meaning [6,7]. They were mainly obtained after some image processing techniques including pattern recognition, optical character recognition, etc.

The main aim of this paper is to apply some of the newest machine learning techniques, actually a capsule deep neural network, for automatic recognition of the historical graffiti, actually letters (XI-XVIII centuries) carved on the stoned walls of St.Sophia cathedral in Kyiv (Ukraine) and estimate the efficiency of capsule networks for the complex geometry, barely discernible shape, and small dataset. The section *II.State of the Art* contains the short characterization of the context, some previous methods used and the results obtained. The section *III.Datasets and Model* describes the datasets and a capsule network applied for their characterization. The section *IV.Results of Machine Learning for Automatic Recognition of Graffiti* contains results of application of capsule network model to the problem. The section V.Discussion and Future Work dedicated to discussion of the results obtained and lessons learned.

## II. STATE OF THE ART

The most part of current and previous researches dedicated to handwriting recognition methods were mainly targeted to various pen, pencil, stilus, or finger writing. The well-known high values of recognition accuracy (>99%) were demonstrated on the MNIST dataset [8] of handwritten digits by a convolutional neural networks (CNNs) [9]. But stone carved handwriting (like shown in Fig.1) has usually much worse quality, distorted and uncompleted state to get the similar values of accuracy. At the moment, most work on character recognition has targeted on pen-on-paper like systems [10]. In many cases the significant data preprocessing procedures are necessary, because without them accuracy can drastically decrease. It is especially important for analysis and recognition of the wood or stone carved letters like historical graffiti which have worse quality, distorted and uncompleted state. For the high-quality preprocessing a priori knowledge about entire glyph is needed. At the moment the Glagolitic and Cyrillic glyph datasets are not available as open source databases except for some cases of their publications [3,4]. But the recent progress of computer vision and machine learning methods allow to apply some of them to improve the current recognition, identification, localization, semantic segmentation, and interpretation of historical graffiti of various origin from different regions and cultures, including Europe (ancient Ukrainian graffiti from Kyivan Rus) [3], Middle East and Africa (Safaitic graffiti) [2], Asia (Chinese hieroglyphs) [11], etc.

Recently, the new image dataset of the carved Glagolitic and Cyrillic letters (CGCL) was assembled and pre-processed for recognition and prediction by machine learning methods. The dataset consisted of more than 4000 images for 34 types of letters. The explanatory data analysis of CGCL and notMNIST datasets shown that the carved letters can hardly be differentiated by dimensionality reduction methods, for example, by t-distributed stochastic neighbor embedding (tSNE) due to the worse letter representation by stone carving in comparison to hand writing. The multinomial logistic regression (MLR) and a 2D convolutional neural network (CNN) models were applied. The MLR model demonstrated the area under curve (AUC) values for receiver operating characteristic (ROC) are not lower than 0.92 and 0.60 for notMNIST and CGCL, respectively. The CNN model gave AUC values close to 0.99 for both notMNIST and CGCL (despite the much smaller size and quality of CGCL in comparison to notMNIST) under condition of the high lossy data augmentation [12]. However, the recent progress of the new types of deep neural networks and, especially, appearance of the new class of capsule networks [13] arise the question about their application to such datasets, especially in the view of the high efficiency of the capsule networks for intersected and distorted objects like handwriting and especially graffiti.

## III. DATASETS AND MODEL

At the moment, more than 7000 graffiti of St. Sophia Cathedral of Kyiv were detected, studied, described, preprocessed, and classified (Fig.1) [14-16]. The most unique corpus of epigraphic monuments of St. Sophia Cathedral of Kyiv (Ukraine) belongs to the oldest inscriptions, which are the most valuable and reliable source to determine the time of construction of the main temple of Kyivan Rus. For example, they contain the cathedral inscriptions-graffiti dated back to 1018–1022, which reliably confirmed the foundation of the St. Sophia Cathedral in 1011. A new image dataset of these carved Glagolitic and Cyrillic letters (CGCL) from graffiti of St. Sophia Cathedral of Kyiv was created to provide glyphs (Fig.2a) for recognition and prediction by multinomial logistic regression and deep neural network [12, 17]. At the moment the whole dataset consists of more than 4000 images for 34 types of letters (classes), but it is permanently enlarged by the fresh contributions [12].

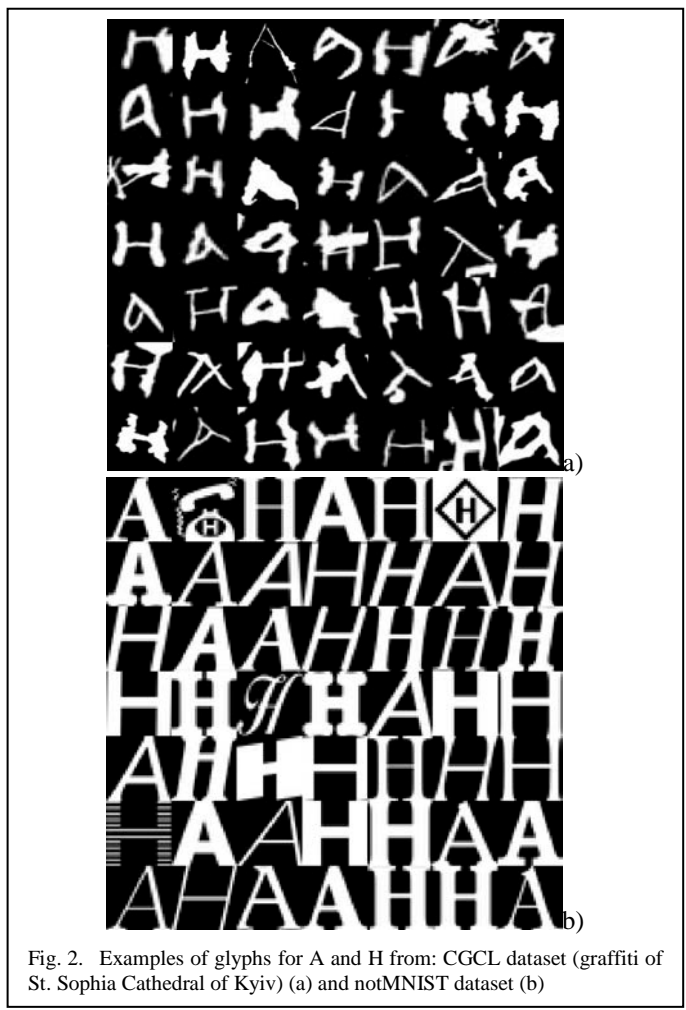

Fig. 2. Examples of glyphs for A and H from: CGCL dataset (graffiti of St. Sophia Cathedral of Kyiv) (a) and notMNIST dataset (b)

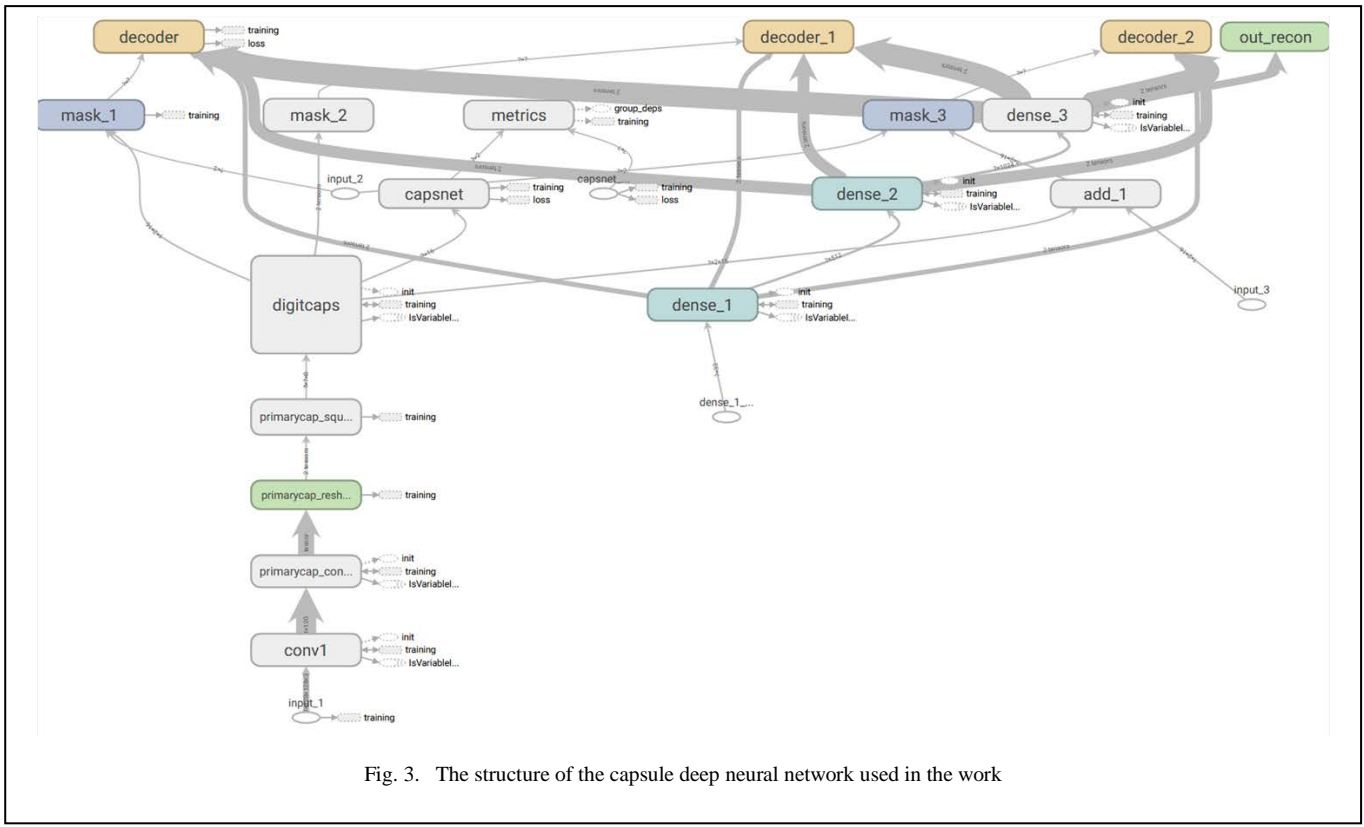

Fig. 3. The structure of the capsule deep neural network used in the work

The second well-known dataset, notMNIST, contains some publicly available fonts for 10 classes (letters A-J) taken from different fonts and extracted glyphs from them to make a dataset similar to MNIST [8]. notMNIST dataset consists of small (cleaned) part, about 19k instances (Fig.2b), and large (uncleaned) dataset, 500k instances [18]. It was used for comparison of the results obtained with GCCL dataset.

As a model, the capsule network model is used which was proposed recently to model hierarchical relationships [13]. In addition to convolutional layers they contain structures (capsules) and allow to reuse output from several of those capsules to form more stable representations (with respect to various lossless and lossy transformations) for higher order capsules. The outputs are vectors that contain probabilities of observations and positions for these observations, which are similar to results of classification with localization tasks performed by some CNNs.

The capsule network (please, see its structure in Fig.3) was applied for the small subsets of GCCL dataset and notMNIST dataset in the following regimes: without data augmentation, with lossless data augmentation, and lossy data augmentation. The capsule network had convolutional and dense layers with capsules with >80 mln. trainable parameters, rectified linear unit (ReLU) and sigmoid activation functions, a binary cross-entropy as a loss function, and Adam (Adaptive Moment Estimation) optimizer with a learning rate of $10^{-4}$.

The subsets contained images of letter A and H (Fig. 2): 180 images for testing, 40 images for validation, and 70 images for testing (for each dataset). The purpose was to check the efficiency of capsule networks for small dataset in the view of various transformations that are typical for the very concise and imperfect handwriting like real graffiti in GCCL dataset. These results are compared here with the similar operations on the other standard dataset that contain similar letters (other than MNIST digit dataset used in numerous other works), for example, on notMNIST dataset.

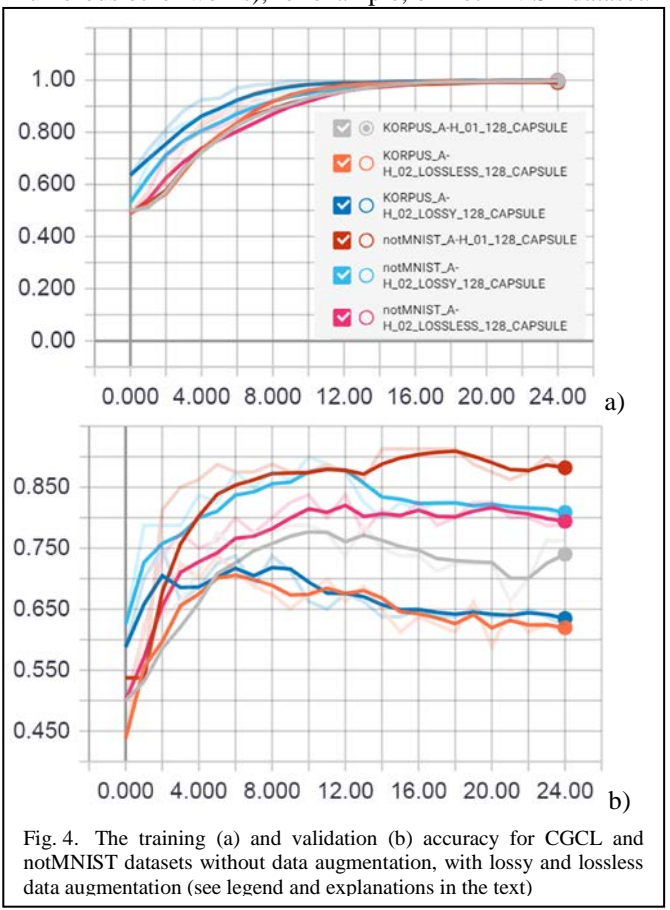

Fig. 4. The training (a) and validation (b) accuracy for CGCL and notMNIST datasets without data augmentation, with lossy and lossless data augmentation (see legend and explanations in the text)

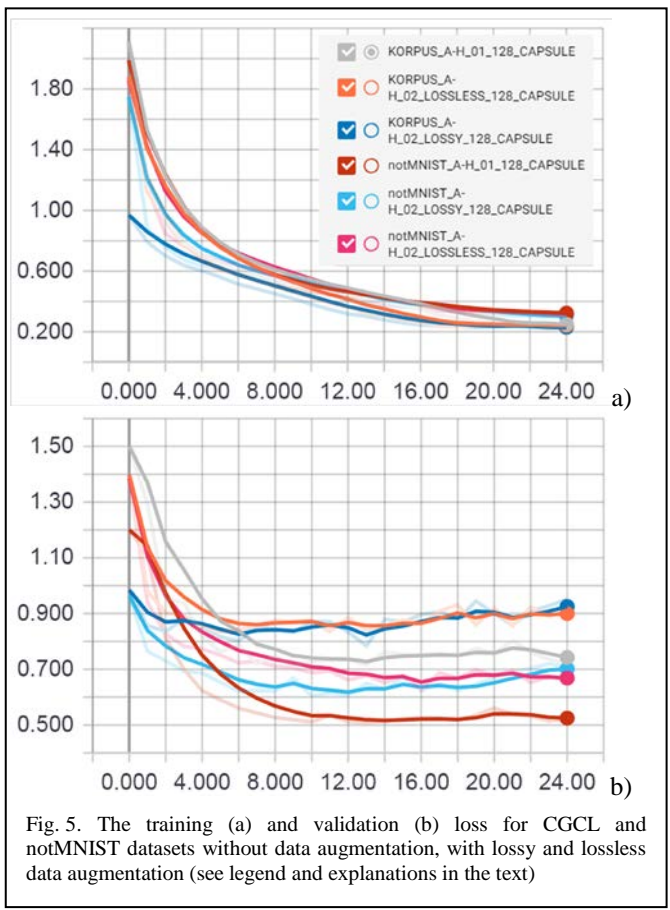

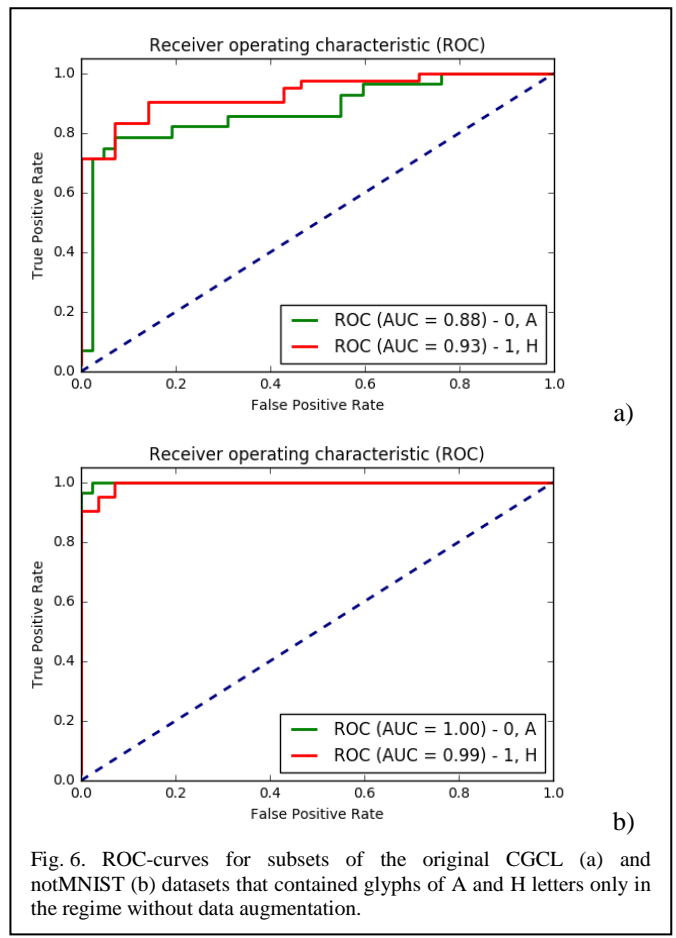

Fig. 5. The training (a) and validation (b) loss for CGCL and notMNIST datasets without data augmentation, with lossy and lossless data augmentation (see legend and explanations in the text).

Fig. 6. ROC-curves for subsets of the original CGCL (a) and notMNIST (b) datasets that contained glyphs of A and H letters only in the regime without data augmentation.

## IV. RESULTS OF MACHINE LEARNING FOR AUTOMATIC RECOGNITION OF GRAFFITI

Despite the much worse quality of CGCL dataset and extremely low number of samples (in comparison to notMNIST dataset) the capsule network model demonstrated much better results than the previously used convolutional neural network (CNN). In Fig.4-5 accuracy and loss results of training and validation attempts are shown for subsets of the CGCL dataset (Fig.4a) and notMNIST dataset (Fig.4b) that contained glyphs of A and H letters only. In contrast to CNN [12] the capsule model does not become overtrained for CGCL dataset (the curves with KORPUS keyword in the legend) and notMNIST dataset (the curves with notMNIST keyword in the legend). The training rate for capsule network model was 5-6 times higher than for CNN [12].

In the regime without data augmentation the validation accuracy and validation loss values demonstrated much lower overtraining for capsule network model (Fig.4.-5) than for CNN model [12].

The lossless data augmentation with addition of the random horizontal and vertical flips of the original images was applied to avoid overtraining (the curves with LOSSLESS keyword in the legend to Fig.4-5). The decrease of accuracy and increase of loss in the lossless data augmentation regime were not so different from the results of the previous regime without data augmentation: ~0.1 for accuracy and ~0.075 for loss.

The lossy data augmentation that included the following transformations: random rotations (up to 40°), width shifts (up to 0.02), height shifts (0.02), shear (0.02), and zoom (0.02) did not improve significantly both datasets accuracy and loss (the curves with LOSSY keyword in the legend to Fig.4-5) in comparison with lossless data augmentation.

These results prove the previously made statements about capsule-like networks [13] that they can be effectively used to better model hierarchical relationships and detect objects that can transform to each other by rotations and translations.

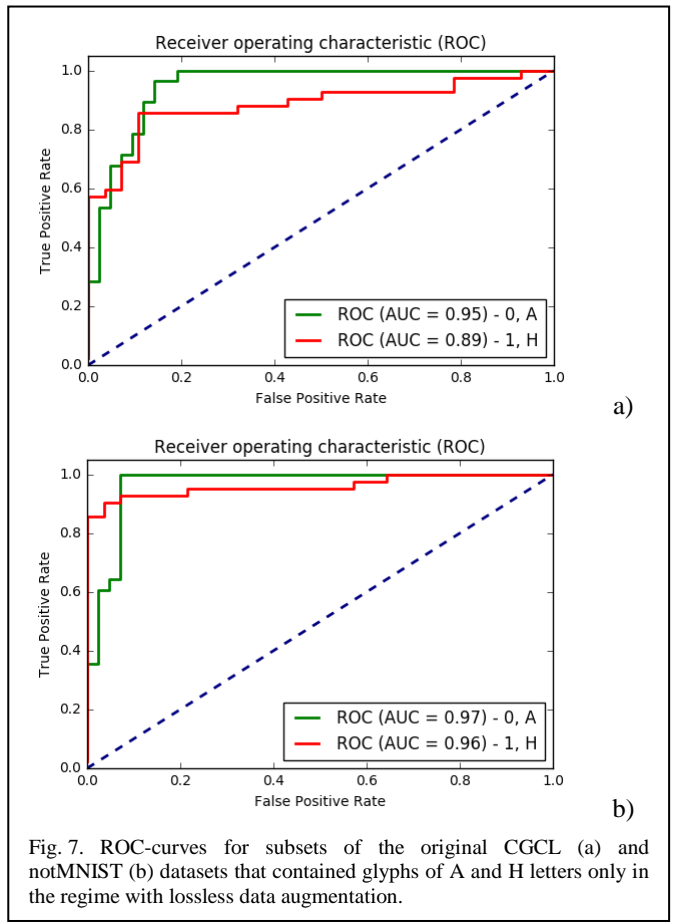

Fig. 7. ROC-curves for subsets of the original CGCL (a) and notMNIST (b) datasets that contained glyphs of A and H letters only in the regime with lossless data augmentation.

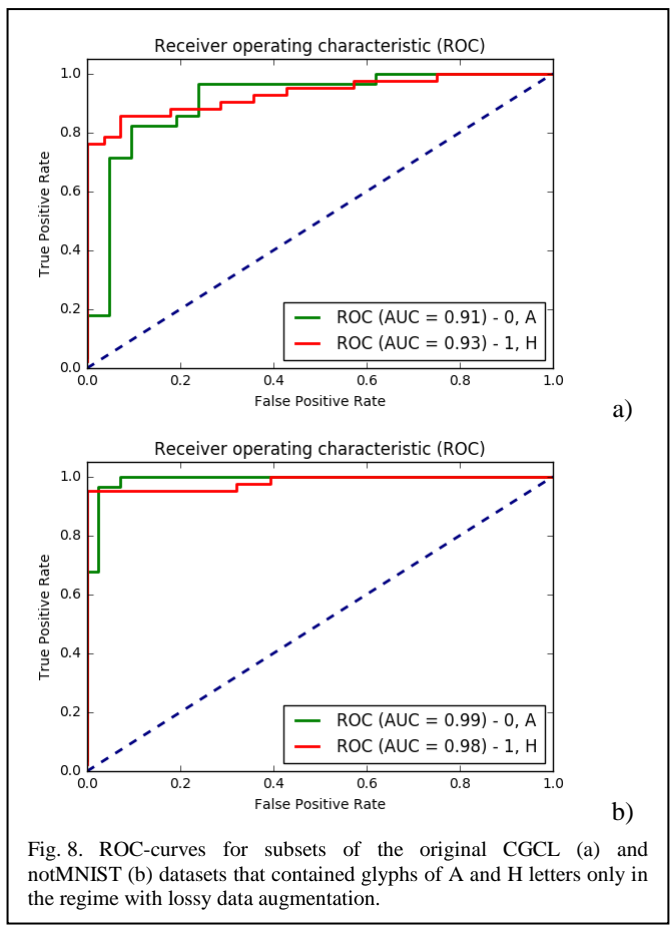

Fig. 8. ROC-curves for subsets of the original CGCL (a) and notMNIST (b) datasets that contained glyphs of A and H letters only in the regime with lossy data augmentation.

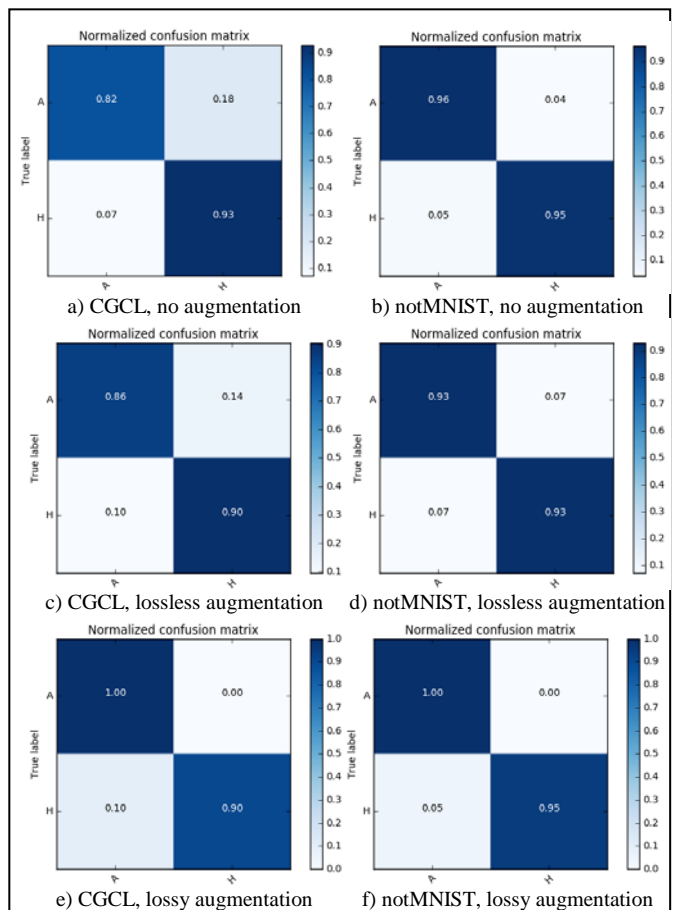

Fig. 8. Confusion matrixes for subsets of the original CGCL (a,c,e) and notMNIST (b,d,f) datasets that contained glyphs of A and H letters only in the regimes: without data augmentation (a,b), with lossless (c,d) and lossy (e,f) data augmentation. The vertical axis corresponds to the true values, and the horizontal one does to the predicted values

The prediction results for test subset of 70 images obtained by the capsule network (Fig.6-8) were also much better than ones obtained by CNN model [12]. The area under curve (AUC) values for receiver operating characteristic (ROC) were higher for the capsule network model than for CNN model: 0.88-0.93 (capsule network, Fig.6a) and 0.50 (CNN) [12] without data augmentation, 0.91-0.95 (capsule network, Fig.7a) and 0.51 (CNN) [12] with lossless data augmentation, and similar results of 0.91-0.93 (capsule network, Fig.8a) and 0.9 (CNN) [12] in the regime of lossless data augmentation only.

The confusion matrixes were much better for capsule network than for CNN model and gave the much lower type I (false positive) and type II (false negative) values in all three regimes of data augmentation (Fig.8). In contrast to the capsule network the CNN model [12] became overtrained for CGCL and notMNIST datasets with the mediocre AOC values for ROC-curve (~0.5) for all regimes except for lossy data augmentation (~0.9).

## V. DISCUSSION AND FUTURE WORK

The results of these classification tasks by machine learning techniques on the basis of the capsule network applied to CGCL and notMNIST datasets shown that the carved letters from CGCL can be easily differentiated by capsule-like networks despite the worse letter representation by stone carving in comparison to glyphs of handwritten and printed letters like notMNIST. The results obtained are good enough for the small dataset with 180 images for training even. In contrast, the CNN models gave the very high AUC values close to 0.9 for both notMNIST and CGCL under condition of the high lossy data augmentation only [12]. These results supports the previous claims that capsule-like networks allow to reduce error rates not only on MNIST digit dataset [12], but on the other notMNIST letter dataset and the more complex CGCL handwriting graffiti letter dataset also. Moreover, application of capsule-like networks allow to reduce training set sizes to 180 images even like in this work. In addition the results obtained by the capsule-like network are considerably better than ones obtained by CNN [12] on the highly distorted (by rotations and translations) and incomplete letters even like CGCL handwriting graffiti. Actually, it is possible by forcing the capsule network model to learn the feature variations in a capsule and extrapolate the possible variations more effectively with less volume of training data. Also it allows to recognize objects regardless of the perspective from which they are viewed and get the high predictions without expensive data augmentation.

The results of this work should be extended by the future research of the other conceptual advantages of capsule network models over CNN models including: viewpoint invariance (recognition of objects regardless of the perspective from which they are viewed), usage of fewer parameters (because connections between layers require fewer parameters), better generalization to new viewpoints, resistance to adversarial attacks.

## VI. CONCLUSIONS

The results supports the assumptions that capsule-like networks allow to reduce error rates not only on digit datasets (like MNIST), but on the other letter datasets (like notMNIST) and the more complex handwriting graffiti letter dataset (like CGCL) also. Moreover, capsule-like networks allow to make precise predictions with much smaller training set sizes like in in this work and they are considerably better than CNNs on the highly distorted and incomplete letters. These results open the new opportunities for development and usage of many more alphabets and writing systems [23-24], and also methods of their automatic recognition for various use cases, for example, touchpads, mouse and body gestures, body sign and eye tracking cameras. This is especially important for health care and elderly care applications [22] on the basis of the newly available information and communication technologies using multimodal interaction by human-computer interfaces like wearable computing, brain-computing interfaces [25], etc. For this purpose the presented capsule network models should be significantly improved to learn the other than spatial relationships between a part and a whole, allowing the latter to be recognized by such relationships. In such case the capsule network models could be sensitive to many additional aspects of complex multimodal interaction data like date, language, authorship, genuineness, and meaning of writing and handwriting, especially like graffiti. But to reach this aim the larger scale and variety datasets and additional research of capsule network models will be necessary. In this sense, the better progress can be provided by developing, sharing, and improving the similar datasets on multimodal interaction in the paradigm of open science, volunteer data collection, processing and computing [2,21].

ACKNOWLEDGMENT

The glyphs of letters from the graffiti [3] were prepared by students and teachers of National Technical University of Ukraine "Igor Sikorsky Kyiv Polytechnic Institute" and can be used as an open science dataset under CC BY-NC-SA 4.0 license (https://www.kaggle.com/yoctoman/graffiti-st-sophia-cathedral-kyiv).

REFERENCES

[1] J. Ancelet, The history of graffiti, University of Central London, 2006.
[2] D. Burt, The online corpus of the inscriptions from ancient north arabia (OCIANA), 2017.
[3] N. Nikitenko,, V. Kornienko, Drevneishie Graffiti Sofiiskogo Sobora v Kieve i Vremya Ego Sozdaniya (Old Graffiti in the St. Sofia Cathedral in Kiev and Time of Its Creation), Mykhailo Hrushevsky Institute of Ukrainian Archeography and Source Studies, Kiev (in Russian) 2012.
[4] S.A. Vysotskii,Drevnerusskie Nadpisi Sofii Kievskoi XI—XIV vv. (Old Russian Inscriptions in the St. Sofia Cathedral in Kiev, 11th—14th Centuries) Kiev: Naukova dumka (in Russian), 1966.
[5] T. Nazarenko, "East slavic visual writing: the inception of tradition," Canadian Slavonic Papers, 43(2-3), 209-225, 2001.
[6] M. Drobysheva, "The difficulties of reading and interpretation of old rus graffiti (the Inscription Vys. 1 as Example)," Istoriya, 6(6 (39)), 10-20, 2015.
[7] O. Pritsak, "An eleventh-century turkic bilingual (turko-slavic) graffito from the st. sophia cathedral in Kiev," Harvard Ukrainian Studies, 6(2), 152-166, 1982.
[8] Y. LeCun, C. Cortes, and C.J. Burges, MNIST handwritten digit database, AT&T Labs. http://yann.lecun.com/exdb/mnist) 1998.
[9] Y. LeCun, L. Bottou, Y. Bengio, and P. Haffner, "Gradient-based learning applied to document recognition," Proceedings of the IEEE, 86(11):2278–2324, 1998.
[10] L.G. Hafemann, R. Sabourin, and L.S. Oliveira, "Offline handwritten signature verification—literature review," In Seventh International Conference on Image Processing Theory, Tools and Applications, pp. 1-8, IEEE, 2017.
[11] J. Winter, "Preliminary investigations on chinese ink in far eastern paintings," Archaeological Chemistry, 207-225, 1974.
[12] N. Gordienko, P. Gang, Yu. Gordienko, W. Zeng, O. Alienin, O. Rokovyi, and S. Stirenko, "Open Source Dataset and Machine Learning Techniques for Automatic Recognition of Historical Graffiti", In Proc. 25th International Conference on Neural Information Processing (Siem Reap, Cambodia), December 14 - 16, 2018; arXiv preprint arXiv:1808.10862 (2018).
[13] S. Sabour, N. Frosst, and G. E. Hinton, "Dynamic routing between capsules", In Advances in Neural Information Processing Systems, pp. 3856-3866, 2017 (https://github.com/XifengGuo/CapsNet-Keras).
[14] N. Nikitenko, V. Kornienko, "Drevneishie graffiti Sofiiskogo sobora v Kieve i ego datirovka," Byzantinoslavica, 68(1), 205-240 (in Russian), 2010.
[15] V.V. Kornienko, Korpus Hrafiti Sofii Kyivskoi, XI - Pochatok XVIII_st, Chastyny I-III (The Collection of Graffito of St. Sophia of Kyiv, 11th – 17th centuries), Parts I-III, Mykhailo Hrushevsky Institute of Ukrainian Archeography and Source Studies, Kiev (in Ukrainian), 2010-2011.
[16] N. Sìnkevič, V. Kornìenko, "Nowe źródła do historii Kościoła unickiego w Kijowie: graffiti w absydzie głównego ołtarza katedry św. Zofii," Studia Źródłoznawcze, 50, 2012.
[17] Glyphs of Graffiti in St. Sophia Cathedral of Kyiv, (https://www.kaggle.com/yoctoman/graffiti-st-sophia-cathedral-kyiv) 2018.
[18] Y. Bulatov, notMNIST dataset. Google (Books/OCR), Tech. Rep. (http://yaroslavvb. blogspot. it/2011/09/notmnist-dataset.html), 2011.
[19] L.V.D. Maaten, and G. Hinton, "Visualizing data using t-SNE," Journal of machine learning research, 9(Nov), 2579-2605, 2008.
[20] P. Schmidt, Cervix EDA and model selection, (https://www.kaggle.com/philschmidt) 2017.
[21] N. Gordienko, O. Lodygensky, G. Fedak, Yu. Gordienko, "Synergy of volunteer measurements and volunteer computing for effective data collecting, processing, simulating and analyzing on a worldwide scale," In: Proc. 38th Int. Convention on Inf. and Comm. Techn., Electronics and Microelectronics, pp. 193-198, IEEE, Opatija, Croatia, 2015.
[22] Peng Gang, Jiang Hui, S. Stirenko, Yu. Gordienko, T. Shemsedinov, O. Alienin, Yu. Kochura, N. Gordienko, A. Rojbi, J.R. López Benito, E. Artetxe González, "User-driven Intelligent Interface on the Basis of Multimodal Augmented Reality and Brain-Computer Interaction for People with Functional Disabilities," In Future of Information and Communication Conference, pp.322-331, IEEE, Singapore, 2018; arXiv preprint arXiv:1704.05915.
[23] S. Hamotskyi, et al., "Automatized generation of alphabets of symbols," In Federated Conference on Computer Science and Information Systems (FedCSIS), pp. 639-642, IEEE, Prague, Czech Republic, 2017.
[24] S. Hamotskyi, et al., "Generating and Estimating Nonverbal Alphabets for Situated and Multimodal Communications," arXiv preprint arXiv:1712.04314 (2017).
[25] Yu. Gordienko, S. Stirenko, O. Alienin, K. Skala, Z. Soyat, A. Rojbi, J.R. López Benito, E. Artetxe González, U. Lushchyk, L. Sajn, A. Llorente Coto, G. Jervan, "Augmented coaching ecosystem for non-obtrusive adaptive personalized elderly care on the basis of cloud-fog-dew computing paradigm," Proc. IEEE 40th Int. Convention on Information and Communication Technology, Electronics and Microelectronics, pp. 387-392, IEEE, Opatija, Croatia, 2017.